\documentclass[10pt,twocolumn,letterpaper]{article}

\usepackage[pagenumbers]{cvpr}      

\usepackage{times}
\usepackage{epsfig}
\usepackage{graphicx}
\usepackage{amsmath}
\usepackage{amssymb}

\usepackage[font=footnotesize,labelfont=bf]{caption}

\usepackage[belowskip=-5pt,aboveskip=3pt]{caption} 

\usepackage{multirow}
\usepackage{tabularx}
\usepackage{dsfont}
\usepackage{url}
\usepackage{color}
\usepackage{amsthm}
\usepackage[export]{adjustbox}
\usepackage{comment}
\usepackage{flushend}

\usepackage[utf8]{inputenc} 
\usepackage[T1]{fontenc}    
\usepackage{url}            
\usepackage{booktabs}       
\usepackage{amsfonts}       
\usepackage{nicefrac}       
\usepackage{microtype}      
\usepackage{enumitem}

\def\ourmethod{{SHAREL}~}
\definecolor{citecolor}{RGB}{65,105,225}
\usepackage[breaklinks=true,colorlinks,citecolor=citecolor,bookmarks=false]{hyperref}

\usepackage{dsfont}
\def\argmax{\operatornamewithlimits{arg\,max}}

\definecolor{dg}{rgb}{0,0.694,0.298}
\definecolor{purple}{rgb}{0.4,0.176,0.569}
\definecolor{royalblue}{RGB}{65,105,225}
\usepackage{pifont}
\newcommand{\figref}[1]{Fig.~\ref{#1}}
\newcommand{\reqref}[1]{Eq.~\eqref{#1}}
\newcommand{\secref}[1]{Sec.~\ref{#1}}
\newcommand{\tableref}[1]{Table~\ref{#1}}

\makeatletter
\DeclareRobustCommand\onedot{\futurelet\@let@token\@onedot}
\def\@onedot{\ifx\@let@token.\else.\null\fi\xspace}
\def\eg{\emph{e.g}\onedot} 
\def\ie{\emph{i.e}\onedot} 
 
\def\etc{\emph{etc}\onedot} 
\def\wrt{w.r.t\onedot} 
\def\etal{\emph{et al}\onedot}
\makeatother

\definecolor{americanrose}{rgb}{1.0, 0.01, 0.24}

\def\cvprPaperID{3256} 
\def\confName{CVPR}
\def\confYear{2022}

\renewcommand{\paragraph}[1]{\vspace{1.25mm}\noindent\textbf{#1}}

\begin{document}

\title{Benchmarking Shadow Removal for Facial Landmark Detection and Beyond}

\author{Lan Fu\textsuperscript{1},
Qing Guo\textsuperscript{2},
Felix Juefei-Xu\textsuperscript{3},
Hongkai Yu\textsuperscript{4},
Wei Feng\textsuperscript{5},
Yang Liu\textsuperscript{2},
Song Wang\textsuperscript{1}\\~\\
\textsuperscript{1}University of South Carolina, USA,~~ 
\textsuperscript{2}Nanyang Technological University, Singapore \\
\textsuperscript{3}Alibaba Group, USA,~~ 
\textsuperscript{4}Cleveland State University, USA,~~ 
\textsuperscript{5}Tianjin University, China
}

\twocolumn[{%
\renewcommand\twocolumn[1][]{#1}%
\maketitle

\begin{center}
\centering
\includegraphics[width=1.0\textwidth]{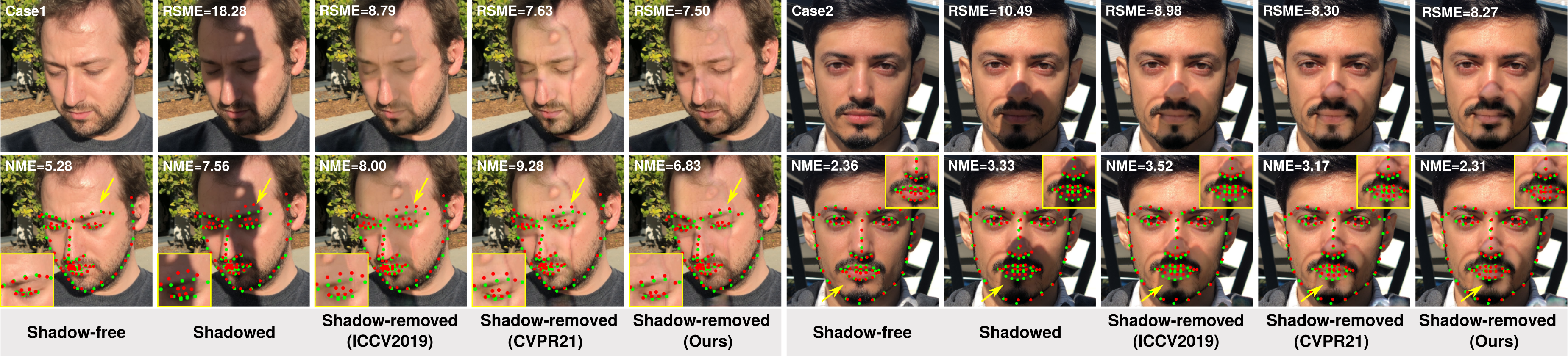}
\captionof{figure}{Shadow removal for facial landmark detection \cite{dong2018style}. \textcolor[RGB]{227,23,13}{Red}: prediction. \textcolor[RGB]{50,180,50}{Green}: ground truth. RMSE measures the shadow removal accuracy, NME evaluates the detection performance. The lower, the better.}
\label{fig:sota_vis}
\vspace{20pt}
\end{center}
}]

\begin{abstract}
Facial landmark detection is a very fundamental and significant vision task with many important applications. In practice, the facial landmark detection can be affected by a lot of natural degradations. One of the most common and important degradations is the shadow caused by light source blocking. While many advanced shadow removal methods have been proposed to recover the image quality in recent years, their effects to facial landmark detection are not well studied. For example, it remains unclear whether the shadow removal could enhance the robustness of facial landmark detection to diverse shadow patterns or not. In this work, for the ﬁrst attempt, we construct a novel benchmark to link two independent but related tasks (\ie, shadow removal and facial landmark detection). 
In particular, the proposed benchmark covers diverse face shadows with different intensities, sizes, shapes, and locations. Moreover, to mine hard shadow patterns against facial landmark detection, we propose a novel method (\ie, adversarial shadow attack), which allows us to construct a challenging subset of the benchmark for a comprehensive analysis.
With the constructed benchmark, we conduct extensive analysis on three state-of-the-art shadow removal methods and three landmark detectors. The observation of this work motivates us to design a novel detection-aware shadow removal framework, which empowers shadow removal to achieve higher restoration quality and enhance the shadow robustness of deployed facial landmark detectors.
\end{abstract}

\section{Introduction}\label{sec:intro}

Facial landmark detection \cite{wu2019facial,zhang2014facial,juefei2013image} is a fundamental step for numerous facial related applications, \eg, face recognition and verification \cite{zhu2015high,liu2018exploring}, 3D face reconstruction \cite{liu2016joint}, and safety-critical applications, \eg, deepfake detection \cite{zhao2021learning, li2020celeb}, and facial reenactment \cite{zhang2020freenet,thies2016face2face} for virtual avatar applications.

While recent deep-learning techniques bring us continuously improved landmark-detection performance, most of them are designed to handle only images of ``clean faces”. However, in real-world applications, face images usually contain image degradations, such as noise, shadow, and haze, which may significantly affect the performance of landmark detectors. Particularly, as a natural phenomenon, shadows are very common on face images – in practice, light to any face region can be occluded by surrounding objects, especially for portrait images captured in the wild. Spatial-variant illumination and color distortion in the shadow region \cite{fu2021auto} degrade the image quality and undermine the image features significantly. As shown in \figref{fig:sota_vis}, shadowed faces hurts the image quality with large root mean square error (RMSE), and presents unreasonable landmark locations for the eyebrows (See Case1) and mouth (See Case2). 

An intuition way to alleviate the performance loss caused by shadow is to restore the underlying shadow-free image utilizing current state-of-the-art (SOTA) shadow removal methods. However, there are two challenges posing to such a solution: \ding{182} The interplay between light, occluder, and the subject directly affects the shadow appearance. As a result, in the real world, shadow patterns are significantly diverse, which increases the difficulty of shadow removal algorithms. \ding{183} Even though shadow removal methods could obtain high visual-quality images with lower RMSE, as shown in \figref{fig:sota_vis}~(Case1), the landmark detection performance even gets worse compared to that of shadow images due to the potential domain shift between landmark detection and image quality enhancement.
All above facts motivate us to answer two basic problems: how shadow affects the landmark detection, and whether shadow removal can benefit the robustness enhancement of landmark detectors.

To this end, for the first attempt, we propose to link the two seemingly independent but intrinsically related tasks, \ie, shadow removal and facial landmark detection, by constructing a totally novel dataset and benchmark. Such a solution has never been tried in both communities before this work. 
Note that, constructing such a benchmark is challenging and not trivial since the shadow patterns are not exhaustive, and existing benchmarking techniques \cite{sagonas2013300,wu2018look} collecting natural images cannot meet the requirements.
To alleviate the challenges, we propose novel solutions to ensure the comprehensiveness: \ding{182} We employ the physical model of shadow and synthesize facial shadow images by considering four common factors (\ie, intensity, size, shape, and location) with three severities,  \ding{183} We think the shadow from the perspective of adversarial attack and propose a totally new attack (\ie, adversarial shadow attack) to identify shadow patterns that are more challenging to landmark detection. \ding{184} We introduce a real-world shadow face dataset for verifying the generalization ability of facial landmark detectors. With these elaborated designs, we are able to quantitatively and systematically study the effect of shadows to the facial landmark detection.

Moreover, we study whether shadow removal can help improve the robustness of landmark detectors covering three SOTA shadow removal methods and three landmark detectors. We observe that shadow removal can not only improve the image visual quality, but also boost the performance of landmark detection – there is a positive correlation between the shadow-removal accuracy and the landmark detection accuracy. 
Note that, such a relationship is not apparent in haze-removal and classification task \cite{pei2019effects} or even opposite in deraining and detection task \cite{hnewa2020object}.
In this work, the relationship is dominate especially when shadow degradation level is higher (\ie, higher-severity shadow and adversarial shadow). It implies that feature embedding spaces of shadow removal and landmark detection aiming to optimize partially overlap with each other, which provides a bridge for the two tasks.
Inspired by this observation, we further propose a new shadow-removal framework regularized by landmark detection to verify whether the two tasks can benefit from each other. As shown in \figref{fig:sota_vis}, the visual quality and landmark detection performance are improved simultaneously.

Overall, we summarize our contributions as follows:
\begin{itemize}[noitemsep, nolistsep,leftmargin=*] 
\item We construct a new shadow-face benchmark \ourmethod, including synthetic shadow-face dataset, adversarial shadow-face dataset, and real shadow-face dataset, by comprehensively considering shadow intensity, size, shape and locations and developing a novel adversarial attack.
\item Based on \ourmethod, we comprehensively and quantitatively study the effect of shadow and shadow removal to image visual quality and the performance of facial landmark detection.
\item We propose a novel shadow removal framework with awareness of facial landmark detection and verify its performance on the proposed benchmark, boosting both the shadow removal and landmark detection performance.
\end{itemize}

\section{Related Work}

\paragraph{Facial landmark detection.}
Deep facial landmark detectors can be classified into two types: direct coordinate regression \cite{toshev2014deeppose,valle2018deeply,li2020structured} and heatmap-based approaches \cite{dong2018style,wang2020deep,zou2019learning}. Coordinate-based landmark detection attempts to locate landmarks directly from images. 
Valle \etal \cite{valle2018deeply} infer landmark locations by a combined network with a tree-structure regression. Heatmap-based methods estimate a likelihood heatmap for each landmark and then infer localization prediction, rendering promising performance over direct regression~\cite{belagiannis2017recurrent}. Dong \etal \cite{dong2018style} propose a style-aggregated network (SAN) to reduce the effect of style variations. Wang \etal \cite{wang2020deep} propose High-Resolution Network (HRNet) to fully explore high resolution information via performing multi-resolution fusion. LUVLi \cite{kumar2020luvli} proposes a deep model to jointly estimate the landmark locations and uncertainty predictions. Graph-based deep learning can also be utilized for facial landmark detection with good robustness and accuracy \cite{li2020structured}.
%

%
\paragraph{Shadow removal.} 
Current state-of-the-art (SOTA) shadow removal methods \cite{qu2017deshadownet,wang2018stacked, zhu2018bidirectional, le2019shadow,hu2019mask,fu2021auto} are divided into two classes: physical shadow decomposition and GAN-based image translation. Physical shadow decomposition performs shadow removal by fusing the shadow image with a relit image \cite{le2019shadow} or fusing multi-exposure images with pixel-wise kernels \cite{fu2021auto} in a paired way. MaskShadow-GAN \cite{hu2019mask} performs shadow removal in an unpaired way by generating a shadow-removed image with the guidance of discriminator. There are two methods to synthesize realistic shadow images. One way \cite{inoue2020learning, zhang2020portrait} is to estimate the shadow parameters based on physical shadow model \cite{shor2008the} with a shadow/shadow-free/binary mask triplet, which is what we used in this work.
The second method \cite{hu2019mask} is to generate shadow images from unpaired shadow-free images and shadow masks utilizing GAN-based image translation, which suffers from artifacts. 

\paragraph{Shadow degradation and landmark detection.}
Shadow, one of the most important and common image degradations, degrades visual quality, resulting in data distribution shift from clean images. Generally, domain gap will lead to performance drop when a pre-trained deep model on clean images is evaluated on degraded domain \cite{sun2018feature,fu2021let}. The effect of shadow on facial landmark detection task is still under-explored. Image-level degradation via shadow can be alleviated by shadow removal such that providing a high-quality image for better visual effect. 
However, visual quality improvement does not always promise performance increasing of a high-level perception task \cite{pei2019effects,hnewa2020object,fu2021let}. Whether shadow removal benefits facial landmark detection remains unexplored. In this work, we firstly attempt to explore the mutual influence of shadow removal and facial landmark detection.

\section{Datasets Construction} \label{sec:dataset_construction}

\subsection{Overview}
Natural shadow presents diverse shadow patterns in the wild due to the influences of occluders and light sources. 
For example, different light occluders can lead to diverse shadow appearances with different sizes and shapes. 
In addition, the illumination level, material of occluders and object surface where shadow casts determine the reflection and scattering of the light, which may affect the intensity at the shadow region.
Nevertheless, enumeration of all permutations formulating patterns is not practical due to dynamic and complex scenes.
To alleviate this issue and analyze the effects of shadow and shadow removal on facial landmark detection extensively, we propose three dataset construction strategies: 
\ding{182} We follow the well-known and widely used physical shadow model to synthesize shadowed faces on the clean facial landmark detection dataset (\ie, 300W \cite{sagonas2013300}) and consider four factors (\ie, intensity, size, shape, and location) with three severities (See \secref{subsec:syn}).
\ding{183} To mine hard shadow images that affect landmark detection easily, we think this problem from the perspective of adversarial attack and propose a novel synthesis method (\ie, \textit{adversarial shadow attack}) in \secref{subsec:adv}.
\ding{184} To address the potential shifting problem between synthesized shadow faces and the real ones, we also introduce 100 real shadow face images as a subset of the whole dataset (See \secref{subsec:real}). We present examples for the three strategies in \figref{fig:dataset_overview} and detail each strategy in the following. 

%

\subsection{Synthetic Shadowed Faces}\label{subsec:syn}
%

\begin{figure*}[t]
\centering
\includegraphics[width=\textwidth]{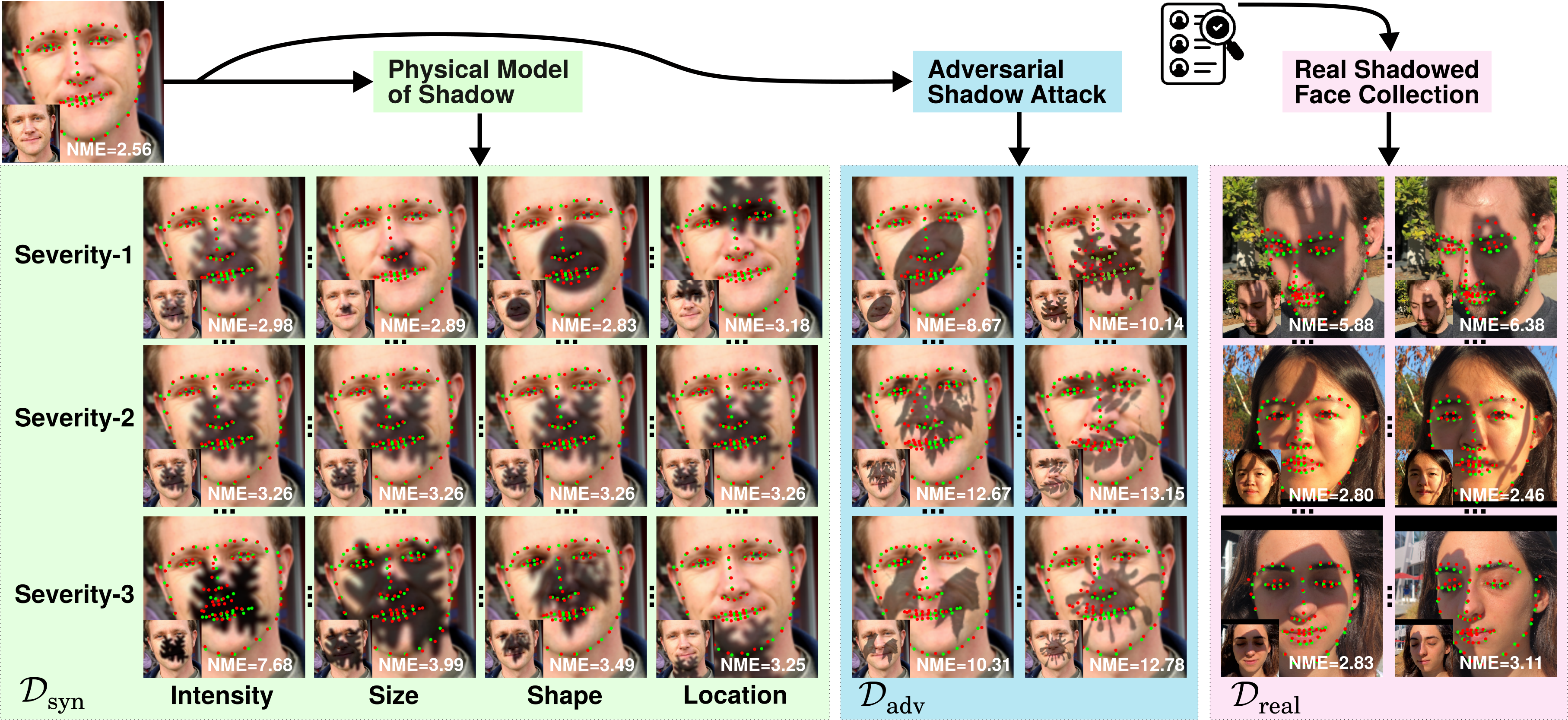}
\caption{Three dataset construction strategies including physical model-based synthesis (See \secref{subsec:syn}), adversarial shadow attack (See \secref{subsec:adv}), and real shadowed face collection (See \secref{subsec:real}). \textcolor[RGB]{50,180,50}{Green}: ground truth. \textcolor[RGB]{227,23,13}{Red}: prediction. NME measures the landmark detection performance. The lower, the better.}
\label{fig:dataset_overview}
\end{figure*}

\paragraph{Physical model of shadow.} 
We adopt the well-known and widely used physical model of shadow in \cite{shor2008the}. Specifically, following the illumination and reflectance formulation of an image \cite{shor2008the}, we can represent a clean (\ie, shadow-free) image captured under a single primary light source as
\begin{align}\label{eq:shadowfree_pixel}
\mathbf{I}_p^\text{cln}= \mathbf{L}_p \mathbf{R}_p = (\mathbf{L}^\mathrm{d}_p+\mathbf{L}^\mathrm{a}_p)\mathbf{R}_p,
\end{align}
where $\mathbf{I}_p^\text{cln}$, $\mathbf{L}_p$, and $\mathbf{R}_p$ are pixel intensity, illumination, and reflectance at the $p$-th pixel, respectively. The illumination stems from two sources, \ie, the direct illumination $\mathbf{L}^\mathrm{d}$ and the ambient illumination $\mathbf{L}^\mathrm{a}$. When an occluder appears in front of the light source, the direct illumination disappears while the ambient illumination is also affected. We can represent the $p$-th shadowed pixel as  
%
\begin{align}\label{eq:shadowed_pxiel}
\mathbf{I}_p^\text{shd}=\alpha\mathbf{L}^\mathrm{a}\mathbf{R}_p = \alpha(\mathbf{I}_p^\text{cln} - \mathbf{L}^\mathrm{d}_p\mathbf{R}_p),
\end{align}
%
where $\alpha$ is a scalar and determines the attenuation of the ambient illumination, which is caused by the occluder. With a clean image $\mathbf{I}^\text{cln}$ and a dark image $\mathbf{I}^\text{shd}$, we can represent an image containing a shadow region as
%
\begin{align}\label{eq:shadowed}
\mathbf{I}=\mathbf{I}^\text{shd}\odot\mathbf{\rho}(\mathbf{D}\odot\mathbf{M})+\mathbf{I}^\text{cln}\odot(1-\mathbf{\rho}(\mathbf{D}\odot\mathbf{M})),
\end{align}
%
where $\mathbf{M}$ is a binary map that defines the shadow region and is determined by the occluder, and $\mathbf{D}$ is a face depth map. 
Note that, the images of living faces have face-like depth information, which are critical for anti-spoofing application. $\mathbf{D}$ can make generated shadow more realistic, which are not considered in previous shadow models \cite{zhang2020portrait,inoue2020learning}. 
Moreover, to generate realistic shadow pattern, we borrow the implementation in  \cite{hanrahan1993reflection,zhang2020portrait} and use function $\rho$ to render the depth-aware mask (\ie $\mathbf{D}\odot\mathbf{M}$) to become a shadow matte image by modeling the light scattering beneath human skin and modeling the spatial variation of the shadow via a spatially-varying blur. Please find more details in \cite{zhang2020portrait}.

Then, we can substitute \reqref{eq:shadowed_pxiel} into \reqref{eq:shadowed} and get
%
\begin{align}\label{eq:shadowedv2}
\mathbf{I} & =  \text{~Shadow}(\mathbf{I}^\text{cln}, \mathbf{M}, \alpha) \nonumber\\
& = (1-(1-\alpha)\mathbf{\rho}(\mathbf{D}\odot\mathbf{M}))\mathbf{I}^\text{cln}
+\alpha\beta\mathbf{\rho}(\mathbf{D}\odot\mathbf{M}),
\end{align}
%
where $\beta=-\mathbf{L}^\mathrm{d}\mathbf{R}$
representing the response of the camera to the reflected direct illumination and the ambient attenuation $\alpha$ does not depend on the light source (\eg, wavelength) \cite{shor2008the}. 
Moreover, as demonstrated and discussed in \cite{inoue2020learning}, $\beta$ is a three-channel vector and can be estimated from the $\alpha$ via a linear transformation. 

Overall, \textit{given a clean face image} $\mathbf{I}^\text{cln}$\textit{, a shadow map $\mathbf{M}$, a depth map $\mathbf{D}$, and the $\alpha$, we can synthesize a shadowed face $\mathbf{I}$.} In practice, we use the 3DDFA-V2 \cite{guo2020towards} to predict the depth map from the clean image.

\paragraph{Synthesized shadows with different factors and severities.}
To cover extensive shadow patterns in the real world, we generate shadowed faces for a clean face image from four factors: intensity, size, shape, and location. 

\begin{enumerate}[itemsep = 0 pt, parsep = 0pt, topsep =0pt,fullwidth,itemindent=1em,label=\roman*.]

\item{\textit{Intensity.}} 
The illumination level and material of object surfaces determine the reflection and scattering of light, resulting in shadow with diverse intensities.
We model the shadow intensity via the parameter $\alpha$ in \reqref{eq:shadowedv2} since it directly models the relationship between shadowed pixels and illuminated pixels. $\alpha$ is about in range [0.0, 1.0) for realistic shadow scene \cite{inoue2020learning}. We uniformly sample $\alpha$ from ranges $[0.8,1.0)$, $[0.4,0.6)$,  $[0.0,0.2)$, for light, medium and heavy shadows. The lower $\alpha$, the heavier the shadow. For different shadow intensity level design, we want to quantify how much texture and content degradation shadow brings, and how that affects visual quality and landmark detection.
We present three kinds of intensities for the same face in \figref{fig:dataset_overview}.


\item{\textit{Size.}} 
The size of an occluder blocking the light and position of the light source directly affect the area of the shadow (\ie, shadow size). We model shadow size via the number of non-zero pixels in $\mathbf{M}$ in \reqref{eq:shadowedv2} and consider three different severities, \ie, small, medium, and large shadow regions. Intuitively, large-size shadow will degrades image quality more than small-size shadow because face-related information (\eg, structure) becomes less. 
Given a specified shadow shape, we can set the shadow areas (\ie, number of non-zero pixels in $\mathbf{M}$) to take up $10\%\thicksim20\%$ , $45\%\thicksim55\%$, and $80\%\thicksim90\%$ areas of the face images by rescaling the shadow region in $\mathbf{M}$, which corresponds to three severities, \ie, small, medium, and large shadow regions.
We show the three different shadow sizes for the same face in \figref{fig:dataset_overview}. 

\item{\textit{Shape.}} 
Occluders with different 3D geometrical shapes and the lights with different positions relative to the same occluder also affect the shadow shapes.
We represent the shadow shape via the shadow mask in $\mathbf{M}$ in \reqref{eq:shadowedv2}. 
To cover diverse shadow shapes, we collect  a silhouette dataset containing 132 shapes of natural objects, and classify them into three levels by a shape complexity metric defined in \cite{chen2005estimating}, which is denoted as $E$. 
The shape complexity metric considers two aspects during measurement, \ie, the distance distribution of the contour points of a shape to its centroid and the smoothness of the contour. 
Intuitively, if the complexity of a shape is low, the shape may tend to be a circle or has smooth contour. 
We present three shapes for the same face in the \figref{fig:dataset_overview}, their complexity values are 0.04, 0.10, and 0.15 from severity 1 to 3.
With the collected silhouette dataset, we first calculate the shape complexity for each collected shape. Then, we sort all shapes according to the complexity and evenly divide them into three severities, \ie, low, medium, and high complexities.

\item{\textit{Location.}}
We further consider the shadow position in the face image, since the shadow with the same intensity, size, and shape may still have different effects to shadow removal and landmark detection methods. For example, facial landmarks include clues of eyebrows, eyes, nose, jaw, and mouth. Shadow degradation to different parts of the facial structure will help quantitatively recognize the importance of each structural information to landmark detection.
We model the shadow location via the centroid position of the shadow mask in $\mathbf{M}$ and consider three scenarios with the position at top, middle, and bottom of the whole face.
For implementation, we split the whole face into three parts, \ie, top, middle, and bottom regions. Then, we shift the centroid point of shadow mask in $\mathbf{M}$ to the center of the three regions. Specifically, if the center point of the whole face is $(\frac{W}{2}, \frac{H}{2})$ where $H$ and $W$ are the height and width of the face, the center points of the top, middle, and bottom regions are $(\frac{W}{2}, \frac{H}{6})$, $(\frac{W}{2}, \frac{H}{2})$, and $(\frac{W}{2}, \frac{5H}{6})$, respectively.
\end{enumerate}
\paragraph{Synthetic shadowed face subset $\mathcal{D}_\text{syn}$.} With the above synthesis strategies,  given a clean face image, we can generate three shadowed faces for each factor, which corresponds to three severities. To consider the effects of all factors, we have $3^4=81$ shadowed faces across all factors and severities for each clean image.
Then, based on the facial landmark dataset 300W \cite{sagonas2013300} that contains 689 clean face images for testing landmark detectors, we can generate a larger dataset with $81\times 689=55,809$ shadowed images. We present some examples in \figref{fig:dataset_overview}. 
Although the constructed dataset covers diverse shadow patterns, it cannot represent all possible situations, in particular, the hard cases that SOTA landmark detectors cannot address. 
To alleviate this issue, we propose a novel adversarial attack in \secref{subsec:adv} based on the physical model of shadow to mine the hard shadow patterns.

\subsection{Adversarially Shadowed Faces}\label{subsec:adv}
Given an image, adversarial attack is to calculate an imperceptible noise-like perturbation under the guidance of a targeted deep model, and then add it to the image. As a result, the corrupted image can mislead the targeted model easily.
Unlike traditional adversarial attacks based on additive perturbations, recently there is a growing trend in developing non-additive adversarial attacks that enjoy better transferability and stealthiness such as blur-based adversarial attacks \cite{neurips20_abba,iccv21_advmot,arxiv21_advbokeh}, attacks based on weather elements \cite{zhai2020s,arxiv21_advhaze} and lighting conditions \cite{arxiv21_ara,ijcai21_ava,icme21_xray,cheng2020adversarial}, and other modalities \cite{gao2020making,tmm21_pasadena,acmmm20_amora,eccv20_spark,iccv21_flat}, \etc. 
We can regard the adversarial attack as a way to mine hard noise patterns that cannot be addressed by the targeted deep model.
Here, we propose a novel attack method, \ie, \textit{adversarial shadow attack}, and further extend it to generate hard shadow patterns that are able to fool the landmark detectors. 
Therefore, we can evaluate the shadow robustness.

Intuitively, we can tune the physical parameters like the $\alpha$ and $\mathbf{M}$ under the supervision of landmark detectors to cover different shadow patterns with different intensities, sizes, shapes, and locations.
Specifically, given a clean face image $\mathbf{I}^\text{cln}$ and a pre-trained landmark detector $\varphi(\cdot)$ we want to evaluate, we can first use \reqref{eq:shadowedv2} to synthesize the shadowed image and feed it to $\varphi(\cdot)$. Then, we get the detection results and calculate the loss according to the ground truth (\ie, $\mathbf{y}$). Our goal is to maximize the landmark detection loss by tuning $\mathbf{M}$ and $\alpha$. We can formulate above process by
%
\begin{align}\label{eq:adv_obj}
\argmax_{\mathbf{M},\alpha,\vartheta} &\mathcal{J}(\varphi(\text{Shadow}(\mathbf{I}^\text{cln}, \text{Aff}_\vartheta(\mathbf{M}),\alpha)), \mathbf{y}), \nonumber\\
&\text{subject~to~~}\|\mathbf{M}-\mathbf{M}_0\|_\text{p}<\epsilon_\text{M}, \nonumber \\
&\|\alpha-\alpha_0\|<\epsilon_{\alpha}, \|\vartheta-\vartheta_0\|_\text{p}<\epsilon_{\vartheta},
\end{align}
%
where $\mathcal{J}(\cdot)$ is the loss function of landmark detection. Note that, different from the raw synthesis function in \reqref{eq:shadowedv2}, we conduct the affine transformation (\ie, $\text{Aff}_{\vartheta}(\cdot)$) on $\mathbf{M}$ before feeding it for synthesis, which allows us to mine more shadow shapes with a given shadow mask. The $\vartheta$ contains six affine parameters. Like general adversarial attack methods, we set the $L_\text{p}$ norm to $\mathbf{M}$, $\alpha$, and $\vartheta$ to force the optimization space within a ball of $\epsilon_\text{M}$, $\epsilon_{\alpha}$, and $\epsilon_{\vartheta}$, around their initialization (\ie, $\mathbf{M}_0$, $\alpha_0$, and $\vartheta_0$), respectively.

To solve the \reqref{eq:adv_obj}, we follow the general adversarial attack methods: \ding{182} We set $\mathbf{M}_0$, $\alpha_0$, and $\vartheta_0$, and get the initial synthesized image. \ding{183} We feed the generated image to the landmark detector  $\varphi(\cdot)$ and calculate the loss. \ding{184} We conduct back-propagation and get the gradients of $\mathbf{M}$, $\alpha$, and $\vartheta$ \wrt the loss function. \ding{185} We calculate the sign of the gradients and use them to update the three variables by multiplying the gradients with three step sizes. \ding{186} We generate a new synthesized image and loop step-2 to step-4 for a number of iterations.
In terms of the initialization, we select $\mathbf{M}_0$ from the collected 132 silhouette images and set $\alpha_0$ to be $0.8$. Then, We initialize $\vartheta_0$ as 
%
$\begin{bmatrix}
1.0 & 0.0 & 0.0\\
0.0 & 1.0 & 0.0
\end{bmatrix}$ 
, $\text{Aff}_\vartheta(\mathbf{M})=\mathbf{M}$ during initialization. 
In terms of the implementation, we set the step size of $\alpha$, $\vartheta$, and $\textbf{M}$ as 0.01, 0.02, and 0.0012, respectively. 
The number of iterations is set to be 40. we use $\infty$ norm for $L_\text{p}$, and set $\epsilon_{\alpha}$, $\epsilon_{\vartheta}$, and $\epsilon_\text{M}$ as 0.4, 0.8, and 0.0.048, respectively. As a result, adversarial shadow images can present more hard shadow patterns against landmark detectors, as shown in \figref{fig:dataset_overview}, the NMEs in $\mathcal{D}_\text{adv}$ could be over 10 compared to around 3 in $\mathcal{D}_\text{syn}$.

\paragraph{Adversarially shadowed face subset $\mathcal{D}_\text{adv}$.} With the above method, given a landmark detector and the 300W dataset, we first conduct attack for each image, and then evaluate the detector on the adversarially shadowed faces. Thus, for each detector, we have an exclusive new version of 689 adversarially shadowed face images to evaluate their robustness.

\subsection{Real Shadowed Faces}\label{subsec:real}

\paragraph{Real shadowed face subset $\mathcal{D}_\text{real}$.} To verify the shadow effect on visual quality and landmark detection performance in the real-world scenario, we introduce a real-world shadow portrait dataset \cite{zhang2020portrait}.
However, this dataset lacks facial landmark annotations for landmark detection evaluation. We first obtain pseudo ground truth by a SOTA pre-trained HRNet \cite{wang2020deep}, and then refine it manually as the final landmark ground truth.
Finally, we have 9 subjects and 100 pairs of shadowed and shadow-free portrait images captured in the outdoor scenes with varied face poses, shadow shapes, and illumination conditions. \figref{fig:dataset_overview} presents some examples.


\section{Shadow Removal \& Landmark Detection Benchmark (SHAREL)}
%

\subsection{Setups} \label{subsec:dataset}

\paragraph{Datasets.} 
As introduced in \secref{sec:dataset_construction}, our main data is constructed based on the landmark detection benchmark 300W~\cite{sagonas2013300}.
300W contains $3,148$ clean face images for training and $689$ clean images for testing. 
Each image is labeled with 68 landmarks.
We construct \ourmethod based on the testing dataset of 300W. As listed in \secref{sec:dataset_construction}, we add shadow patterns to the 300W and get $\mathcal{D}_\text{syn}$; we propose adversarial shadow attack and obtain $\mathcal{D}_\text{adv}$ for each landmark detector; we collect real shadowed faces (\ie, $\mathcal{D}_\text{real}$) to further enrich our dataset.
Finally, our dataset $\{\mathcal{D}_\text{syn}; \mathcal{D}_\text{adv}; \mathcal{D}_\text{real}\}$ has 
$\{55,809; 689; 100\}$ shadowed and shadow-free image pairs (total $56,598$ pairs) that are labeled with 68 landmarks.

We additionally construct $\{\mathcal{D}^\text{t}_\text{syn}, \mathcal{D}^\text{t}_\text{adv}\}$ from a randomly selected subset (1,500 clean images) of 300W training set for training shadow removal models. Each of $\{\mathcal{D}^\text{t}_\text{syn}, \mathcal{D}^\text{t}_\text{adv}\}$ contains 1,500 shadow-free and shadowed image pairs. For creating $\mathcal{D}^\text{t}_\text{syn}$, each clean image uniformly selects a severity for each factor to generate the shadow image. $\mathcal{D}^\text{t}_\text{adv}$
follows the same shadow generation way of $\mathcal{D}_\text{adv}$.

\paragraph{Metrics.}
To clarify the shadow and deshadow effect on image quality, we adopt the Root Mean Square Error (RMSE) metric in LAB color space for evaluation, similar to \cite{fu2021auto,le2019shadow,hu2019mask}. 
For facial landmark detection evaluation, we adopt Normalized Mean Error (NME) metric with inter-ocular distance as normalization strategy following \cite{kumar2020luvli,dong2018style,wang2020deep}. Both the lower, the better.
%

\begin{figure*}[t]
\centering
\includegraphics[width=\textwidth]{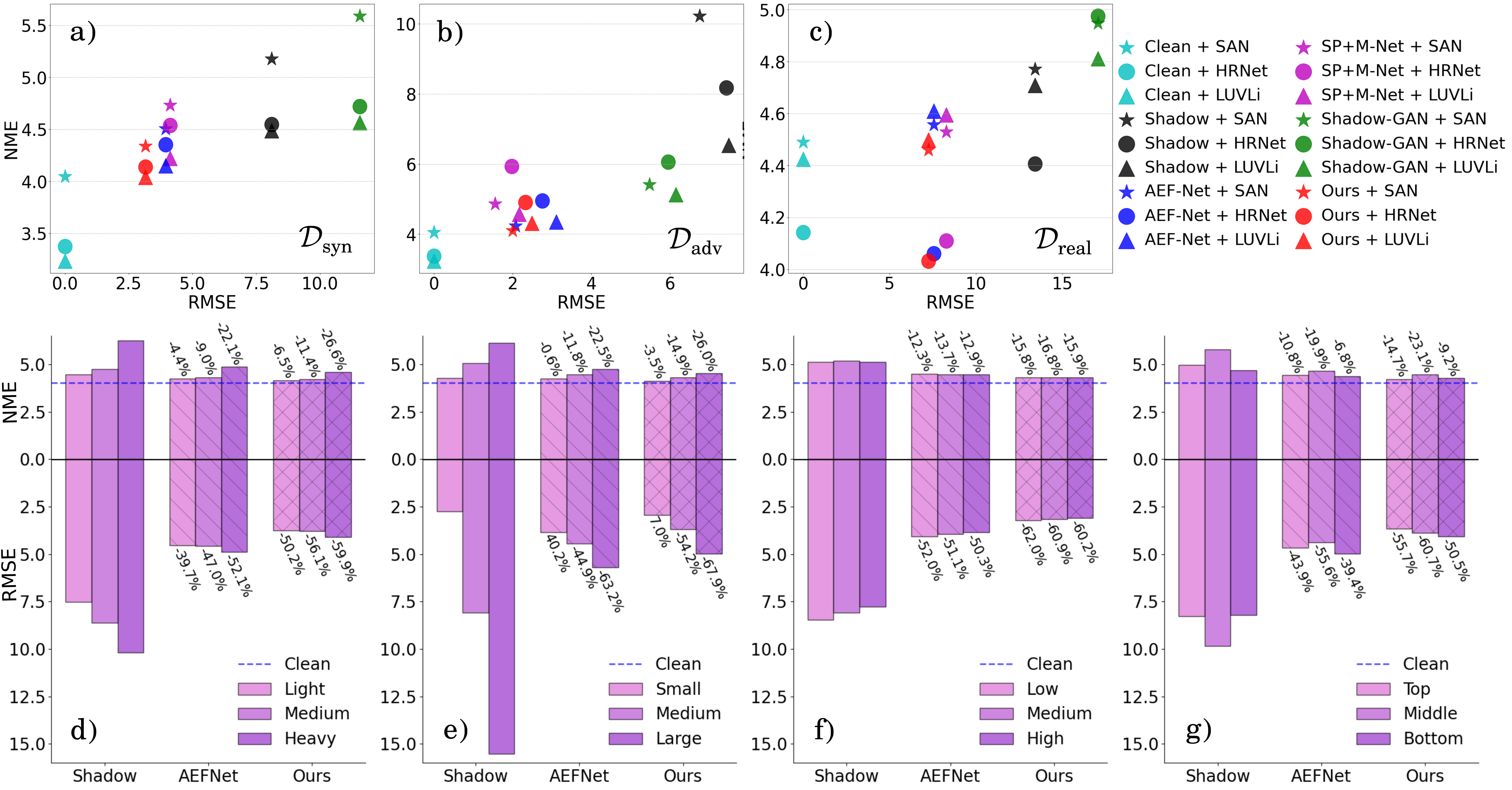}
\caption{Shadow removal and landmark detection performance on \ourmethod.
(a-c): shadow removal (RMSE) and landmark detection (NME) results of $\{\mathcal{D}_\text{syn}, \mathcal{D}_\text{adv}, \mathcal{D}_\text{real}\}$ subsets, respectively. Each color represents results on shadow-free images (\eg, Clean+*), shadow images (\ie, Shadow+*), and shadow-removed images with four shadow removal methods (\eg, AEFNet/SP+M-Net/MaskShadow-GAN/Ours+*). Different icon shapes represent different landmark detectors.
(d-g): shadow pattern analysis of landmark detection (NME) and shadow removal (RMSE) results of $\mathcal{D}_\text{syn}$ for \textbf{intensity} (d), \textbf{size} (e), \textbf{shape} (f), and \textbf{location} (g). Blue dash line represents the result on clean images by the pre-trained landmark detector SAN \cite{dong2018style}.
Each group represents results on shadow images (\ie, Shadow), and shadow-removed images with two shadow removal methods (\eg, AEFNet/Ours). Each color represents a severity type. Relative performance gain, \ie, the percent of NME/RMSE drops, after shadow removal compared to shadow images is listed for AEFNet and Ours.
}
\label{fig:all_rm_ld}
\end{figure*}

\paragraph{Evaluated methods.}
With our \ourmethod, we can evaluate the quality restoration capability of the shadow removal methods and the detection accuracy of facial landmark detectors on different shadow or deshadowed patterns.
We first analyze three SOTA facial landmark detectors, \ie, SAN~\cite{dong2018style}, HRNet~\cite{wang2020deep}, and LUVLi~\cite{kumar2020luvli}, under different shadow patterns. All landmark detectors are pre-trained on clean face images. Further, we utilize three SOTA deep shadow removal methods, \ie, MaskShadow-GAN~\cite{hu2019mask}, SP+M-Net~\cite{le2019shadow}, and AEFNet~\cite{fu2021auto}, to handle the shadowed faces in \ourmethod and discuss whether and how these methods can help improve landmark detection performance. All shadow removal algorithms are trained on dataset $\mathcal{D}^\text{t}_\text{syn}$ and $\mathcal{D}^\text{t}_\text{adv}$ separately for fair comparison, and shadow removal models trained on $\mathcal{D}^\text{t}_\text{syn}$ are also utilized to test on real data.

\subsection{Evaluation Results and Discussion}

\paragraph{Effects of shadow to image quality and facial landmark detection.} In \figref{fig:all_rm_ld}(a-c), we report the RMSEs of shadow images and landmark detection results with NMEs in $\{\mathcal{D}_\text{syn}, \mathcal{D}_\text{adv}, \mathcal{D}_\text{real}\}$ to identify the shadow degradation on image quality and detection performance. \figref{fig:all_rm_ld}(d-g) report the shadow pattern analysis on $\mathcal{D}_\text{syn}$ with four factors. The detector adopted in(d-g) is SAN \cite{dong2018style}. The results show that:
\ding{182} Compared with shadow-free images, shadow images have high RMSEs since the shadow harms the image quality significantly. More intense the shadow degradation, worse the visual quality. For example, the RMSE of shadow and shadow-free images of large-size with 15.52 is higher than that of small-size with 2.74 in $\mathcal{D}_\text{syn}$ (\figref{fig:all_rm_ld}e). Intensity, size, and location, instead of shape, are dominant factors affecting the shadow degradation.
\ding{183} According to the NME results, we observe that: the performance of all landmark detectors drops when shadow appears in images and hard shadow pattern, \ie, higher-severity shadow and adversarial shadow, hurts the detection task most. Specifically, the landmark detector SAN \cite{dong2018style} achieves 4.05 NME on clean images of $\mathcal{D}_\text{adv}$, while the NME of shadow images increases by 152.3\% to 10.22 (\figref{fig:all_rm_ld}b). In $\mathcal{D}_\text{syn}$, heavy-intensity shadow achieves 6.26 NME with 54.7\% performance drop compared to NME of clean images, while the performance loss caused by light-intensity shadow is 10.2\% by SAN \cite{dong2018style} (\figref{fig:all_rm_ld}d). 
%

\textit{In summary, shadow hurts the image quality and landmark detection significantly. Higher-severity presents high degradation capacity, that is, two tasks suffer from larger performance loss with increasing RMSEs and NMEs. The same performance loss trend appears in the image quality and landmark detection.}

\paragraph{Effects of shadow removal to image quality enhancement and facial landmark detection.} We perform shadow removal on shadow images, and present RMSEs and NMEs of shadow-removed images in $\{\mathcal{D}_\text{syn}, \mathcal{D}_\text{adv}, \mathcal{D}_\text{real}\}$ to evaluate the effectiveness of shadow removal methods. The results are shown in the \figref{fig:all_rm_ld}. We can observe that: 
\ding{182} Shadow removal methods present different capabilities on the image quality enhancement (\figref{fig:all_rm_ld}(a-c)). To be specific, SP+M-Net \cite{le2019shadow} and AEFNet \cite{fu2021auto} can enhance the image quality significantly in all subsets. MaskShadow-GAN \cite{hu2019mask} further hurts the quality in the subsets $\{\mathcal{D}_\text{syn}, \mathcal{D}_\text{real}\}$ while achieving counterpart result in $\mathcal{D}_\text{adv}$. The former mainly stems from that MaskShadow-GAN, \ie, a GAN-based image translation method, introduces artifacts during training. The reason why MaskShadow-GAN performs better on $\mathcal{D}_\text{adv}$ may be that shadow pattern generated by MaskShadow-GAN overlaps with that of $\mathcal{D}_\text{adv}$. Specifically, during MaskShadow-GAN training, it also generates diverse shadow patterns taking unpaired shadow-free images and shadow masks as input, and such shadow-pattern images are not covered by normal shadow images via a discriminator in a adversarial training way, similar to the adversarial generation process of $\mathcal{D}_\text{adv}$. 
\ding{183} Higher-severity shadow pattern achieves much larger relative gain for image quality enhancement. 
For example, large-size shadow-removed images acquire 63.2\% visual quality improvement compared to 40.2\% quality degradation of small-size shadow in $\mathcal{D}_\text{syn}$ (\figref{fig:all_rm_ld}e). The latter further quality degradation stems from the over smoothing of current shadow removal methods. In addition, $\mathcal{D}_\text{adv}$ also achieves much larger gain with 69.3\% compared to 51.2\% of $\mathcal{D}_\text{syn}$ by SAN (\figref{fig:all_rm_ld}(a-b)). 
\ding{184} The same performance gain trend of SOTA shadow removal methods and higher-severity shadow pattern presents in the landmark detection evaluation. In \figref{fig:all_rm_ld}e, the large-size shadow pattern obtains the highest 22.5\% NME decreasing compared to 0.6\% of small-size shadow. The $\mathcal{D}_\text{adv}$ achieves 58.5\% performance improvement compared to 13.0\% of $\mathcal{D}_\text{syn}$ by SAN (\figref{fig:all_rm_ld}(a-b)).

\textit{In summary: \ding{182} Current SOTA shadow removal methods can effectively improve the image quality and landmark detection simultaneously. \ding{183} Higher-severity achieves much larger performance gain after shadow removal for image quality and landmark detection. \ding{184} There is a positive correlation between shadow removal and landmark detection tasks. To be specific, landmark detection performance decreases with degraded image quality caused by shadow and improves with increasing image quality after shadow removal. In particular, when image quality suffers from higher degradation, \ie, higher-severity shadow and adversarial shadow, the performance gain trend keeps consistent for the two tasks after shadow removal.} 

Note that, such positive correlation does not always exist in computer vision tasks. For example, deraining even hurts the object detection performance on rainy images \cite{hnewa2020object}. Haze-removal improves classification task with very limited margin \cite{pei2019effects}. In this work, the positive correlation between shadow removal and landmark detection implies that the embedding spaces they optimized somehow overlap with each other. However, previous shadow removal works \cite{hu2019mask,le2019shadow,fu2021auto} only focus on recovering pleasing visual images ignoring the mutual influence between them. 
We propose a novel framework to explore the mutual influence of the two tasks to verify whether they can benefit from each other.


\section{Landmark-regularized Shadow Removal}

\begin{figure}[t]
\centering
\includegraphics[width=1.0\columnwidth]{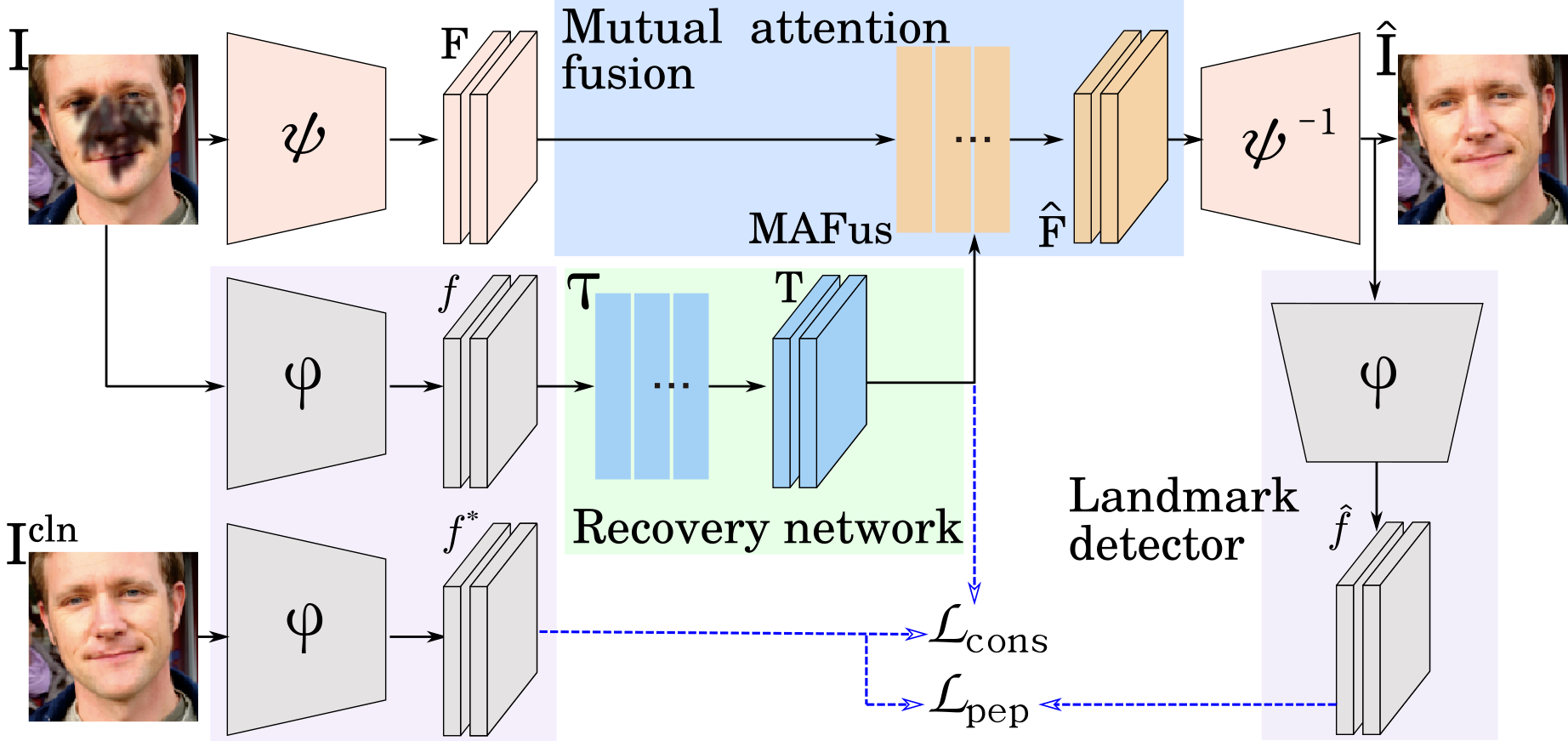}
\caption{Landmark-regularized shadow removal network.}
\label{fig:framework}
\end{figure}

To link the facial landmark detection and shadow removal, we propose to introduce the landmark detection embedding to regularize the shadow removal by the \textit{mutual attention fusion module}. Moreover, we propose extra regularization loss functions by jointly considering the image reconstruction and landmark detection, as shown in \figref{fig:framework}.

Given a shadow image $\mathbf{I}$, a shadow removal method can be generally represented as 
%
\begin{align}\label{eq:shadow_removal}
\hat{\mathbf{I}} = \psi^{-1}(\mathbf{F}), \text{and}~\mathbf{F} = \psi(\mathbf{I}),
\end{align}
where $\psi(\cdot)$ and $\psi^{-1}(\cdot)$ are the encoder and decoder of the shadow removal model, respectively, and $\hat{\mathbf{I}}$ is the shadow-removed image.
Considering a backbone model $\varphi(\cdot)$ of a landmark detector, we propose to fuse the embeddings of the shadow removal model and landmark detection model, and modify \reqref{eq:shadow_removal} as
%
\begin{align}\label{eq:shadow_removal_new}
\hat{\mathbf{I}} = \psi^{-1}(\hat{\mathbf{F}}), \text{and}~\hat{\mathbf{F}} = \text{MAFus}(\psi(\mathbf{I}),\tau(\varphi(\mathbf{I}))),
\end{align}
where $\tau(\cdot)$ is a recovery network to align the embedding of $\varphi(\mathbf{I})$
to the space of $\varphi(\mathbf{I}^\text{cln})$ for better detection-aware embedding. $\text{MAFus}(\cdot)$ is the mutual attention fusion module and can leverage the features from different networks and integrate information complementary.

\paragraph{Mutual attention fusion (MAFus).} 
Inspired by the recent self-mutual attention based on non-local module \cite{wang2018non} for fusing multiple modalities \cite{liu2020learning}, we employ it for embedding fusion. Specifically, we first map the input features $\mathbf{F}=\psi(\mathbf{I})$ and $\mathbf{T}=\tau(\varphi(\mathbf{I}))$ to three spaces like non-local network
%
\begin{align}\label{eq:nonlocal-1}
\theta^\text{f}(\mathbf{F}) & = \mathbf{F}\mathbf{W}_{\theta}^\text{f},
\phi^\text{f}(\mathbf{F}) = \mathbf{F}\mathbf{W}_{\phi}^\text{f},
g^\text{f}(\mathbf{F}) = \mathbf{F}\mathbf{W}_{g}^\text{f}, \\
\theta^\text{t}(\mathbf{T}) & = \mathbf{T}\mathbf{W}_{\theta}^\text{t},
\phi^\text{t}(\mathbf{T}) = \mathbf{T}\mathbf{W}_{\phi}^\text{t},
g^\text{t}(\mathbf{T}) = \mathbf{T}\mathbf{W}_{g}^\text{t}.
\end{align}
%
Then, we can use the Gaussian function for self similarity calculation on $\theta^*$ and $\phi^*$ spaces
%
\begin{align}\label{eq:nonlocal-2}
f^\text{f}(\mathbf{F})=\theta^\text{f}(\mathbf{F})\phi^\text{f}(\mathbf{F})^{\top}, f^\text{t}(\mathbf{T})=\theta^\text{t}(\mathbf{T})\phi^\text{t}(\mathbf{T})^{\top}.
\end{align}
%
After that, we calculate the mutual attention based on the two similarity results
%
\begin{align}\label{eq:nonlocal-3}
\mathbf{A}^\text{f}(f^\text{f}(\mathbf{F}),f^\text{t}(\mathbf{T}))=\text{softmax}(f^\text{f}(\mathbf{F})+\gamma^\text{t}f^\text{t}(\mathbf{T})),\\
\mathbf{A}^\text{t}(f^\text{f}(\mathbf{F}),f^\text{t}(\mathbf{T}))=\text{softmax}(f^\text{t}(\mathbf{F})+\gamma^\text{f}f^\text{f}(\mathbf{T})),
\end{align}
%
where $\gamma^\text{t}$ and $\gamma^\text{f}$ are pixel-wise attention weights to fuse embedding attentions, and are predicted by the concatenation of $\mathbf{F}$ and $\mathbf{T}$.
With the mutual attention result $\mathbf{A}^\text{f}$ and $\mathbf{A}^\text{t}$, we can obtain the non-local outputs of $\mathbf{F}$ and $\mathbf{T}$,
%
\begin{align}\label{eq:nonlocal-4}
\mathbf{Z}^\text{f}=(\mathbf{A}^\text{f}g^\text{f}(\mathbf{F}))\mathbf{W}_z^\text{f}+\mathbf{F},  \\ 
\mathbf{Z}^\text{t}=(\mathbf{A}^\text{t}g^\text{t}(\mathbf{T}))\mathbf{W}_z^\text{t}+\mathbf{T}.
\end{align}
The final output of MAFus is the concatenation of $\mathbf{Z}^\text{f}$ and $\mathbf{Z}^\text{t}$, \ie, $\hat{\mathbf{F}}=[\mathbf{Z}^\text{f}, \mathbf{Z}^\text{t}]=\text{MAFus}(\mathbf{F},\mathbf{T})$.

\paragraph{Loss functions.}
We employ $L_{1}$ distance for the image reconstruction loss $\mathcal{L}_\mathrm{pix}(\hat{\mathbf{I}}, {\mathbf{I}}^\text{cln}) =\|{\mathbf{I}}^\text{cln}-\hat{\mathbf{I}}\|_{1}$. We propose three more regularization loss functions to explore the detection embedding guidance for shadow removal, which are detection regularization loss $\mathcal{L}_\mathrm{det}$, detection-aware perceptual loss $\mathcal{L}_\mathrm{pep}$, and detection-aware consistency loss $\mathcal{L}_\mathrm{cons}$.

Detection regularization loss aims to provide a regularization item for constraining the shadow removal process to satisfy the landmark detection. The weights of pre-trained landmark detector with clean images are fixed and only shadow removal is optimized. 
Given a shadow-removed image $\hat{\text{I}}$ , detection embedding $\hat{f}$ and heatmap $\hat{h}$ can be inferred by 
$\hat{f}, \hat{h} =\varphi(\hat{\mathbf{I}})$. For its corresponding shadow-free image, 
$f^{*}, h^{*}=\varphi(\mathbf{I}^\text{cln})$. The detection regularization loss is defined as 
$
\mathcal{L}_\mathrm{det}(\hat{h}, h^*) = \mathrm{MSE}(\hat{h}, h^*).
$
 Moreover, inspired by perceptual loss \cite{johnson2016perceptual}, we align the detection embeddings of shadow-removed and clean images by 
$
\mathcal{L}_\mathrm{pep}(f^\text{*}, \hat{f}) = \mathrm{MSE}(f^{*}, \hat{f}).
$
 Finally, we propose detection-aware consistency loss $\mathcal{L}_\mathrm{cons}$ to align the embeddings of $\tau(\varphi(\mathbf{I}))$ and $\varphi(\mathbf{I}^\text{cln})$. 
The $\mathcal{L}_\mathrm{cons}$ aims to drive transformed detection embedding of shadow image to that of shadow-free image, their consistency renders the $\tau(\varphi(\mathbf{I}))$ to provide rich complementary information for better shadow removal guidance. It is formulated to $\mathcal{L}_\mathrm{cons}(f^\text{*}, \mathbf{T}) = \text{MSE}(f^\text{*},\mathbf{T})$. The total loss of the proposed framework is
%
\begin{align}
\mathcal{L}= \mathcal{L}_\mathrm{pix} + \lambda_{1}\mathcal{L}_\mathrm{det} + \mathcal{L}_\mathrm{cons} + \lambda_{2} \mathcal{L}_\mathrm{pep},
\end{align}
%
where $\lambda_1$ and $\lambda_2$ are set to 0.1 and 10 in our experiments.

\section{Experiments}

\paragraph{Setups.} 
We conduct extensive experiments to verify our proposed landmark-regularized shadow removal method. 
Based on baseline method AEFNet \cite{fu2021auto}, we cumulatively add each module for contribution evaluation: 1) Detection regularization loss $\mathcal{L}_\mathrm{det}$. We adopt SAN \cite{dong2018style} as the weight-fixed landmark detector. 2) Mutual attention fusion module $\text{MAFus}$. A shadow image is fed into SAN and the output detection feature map will be directly fused with shadow removal feature map via $\text{MAFus}$. 3) Detection-aware consistency loss $\mathcal{L}_\mathrm{cons}$. Detection feature map of shadow image is fed into the recovery network $\tau$ for feature alignment, followed by performing $\text{MAFus}$.
4) Detection-aware perceptual loss $\mathcal{L}_\mathrm{pep}$. Feature shift of shadow-removed image and clean image is further optimized by $\mathcal{L}_\mathrm{pep}$. The training set and testing set are $\mathcal{D}^\text{t}_\text{syn}$ and $\mathcal{D}_\text{syn}$. Results are shown in \tableref{tab:abl}. 

\paragraph{Implementing details.} 
The proposed pipeline is implemented in PyTorch. We build our proposed framework based on shadow-removal method AEFNet \cite{fu2021auto} and training setting keeps the same with official publicized code of AEFNet. 

\begin{table}[ht]
\caption{Ablation study of shadow removal and landmark detection results on the $\mathcal{D}_\text{syn}$ dataset.}
\centering
\footnotesize
\resizebox{0.95\linewidth}{!}{
\begin{tabular}{r|ccc|c}
\toprule
\multirow{2}{*}{Methods} & \multicolumn{3}{c|}{Shadow removal / RMSE} & \multirow{2}{2cm}{Landmark detection / NME}\\
\cline{2-4} 
 & Shadow & Non-shadow & All & \\
\midrule
Clean                                & 0.00  & 0.00 & 0.00 & 4.04\\
w/ shadow                            & 33.27 & 0.53 & 8.09 & 5.17  \\
\midrule
AEFNet~\cite{fu2021auto}             & 9.14 & 2.39 & 3.95 & 4.50\\
+ $\mathcal{L}_\mathrm{det}$         & 7.56 & \color{red}{1.93} & 3.23 & 4.40 \\
+ $\text{MAFus}$                     & 7.16 & 1.96 & 3.16 & 4.35 \\
+ $\mathcal{L}_\mathrm{cons}$         & 7.18 & \color{red}{1.93} & \color{red}{3.14} & 4.34 \\
+ $\mathcal{L}_\mathrm{pep}$         & \color{red}{7.09} & 1.97 & 3.15 & \color{red}{4.33}\\
\bottomrule
\end{tabular}\label{tab:abl}
}
\end{table}

\paragraph{Results and discussion.} 
In Table \ref{tab:abl}, it turns out that: 1) Detection loss regularization improves the shadow removal capacity with 18.2\% decreasing RMSE compared to baseline method in the whole image. Landmark detection performance also benefits from it with 2.2\% decreasing NME. 2) The attention-based feature fusion $\text{MAFus}$ and detection-aware consistency loss further reduce the RMSE to 3.16 and 3.14 in the whole image. Correspondingly, landmark detector also performs better with reaching 4.35 NME and 4.34 NME, respectively. 3) With the detection-aware perceptual loss, the propose method performs best in the shadow region with 7.09 RMSE for image quality evaluation and with 4.33 NME for landmark detection evaluation. Compared to shadow images, proposed method improves the visual quality by 78.7\% in the shadow region, and increases the landmark detection performance by 16.2\%. In \figref{fig:sota_vis}(Case2) and \figref{fig:all_rm_ld}c, our proposed method even achieves better detection performance on shadow-removed images compared to on clean images by SAN. 
%

\section{Conclusion}
We have proposed a shadow-removal benchmark dataset \ourmethod to explore the mutual influence of shadow removal and facial landmark detection tasks. We first proposed three strategies to construct the benchmark. Based on physical shadow model, we synthesize the shadowed faces considering four factors (\ie, intensity, size, shape, and location) with three severities to cover diverse shadow patterns. We also proposed an adversarial shadow attack as hard shadow patterns to make the landmark detection fail easily. Real shadowed face dataset for landmark detection is to reduce the distribution shift with synthetic data. Based on the proposed benchmark, we explored the shadow and shadow-removal effect on visual quality and landmark detection tasks comprehensively. We observed that there is a highly positive correlation between shadow removal and the facial landmark detection task, especially, when degradation level is higher. We then proposed a novel shadow-removal framework regularized by facial landmark detection to benefit each other. We verified the effectiveness of our proposed method in synthetic data, adversarial data and real data. One potential limitation is that this work mainly focuses on image tasks and we will extend it to video tasks in the near future work.


\appendix
\section{Appendices}
\renewcommand\thefigure{\Alph{figure}}  
\renewcommand\thetable{\Alph{table}}  
\renewcommand\thesection{\Alph{section}}  

\subsection{Results under Different Shadow Patterns}
We conduct shadow pattern analysis on $\mathcal{D}_\text{syn}$ considering different factors. When evaluating the shadow and shadow removal effects by one factor to image quality and facial landmark detection, we enumerate other factors for a comprehensive analysis. For example, given the $689$ clean face images, when evaluating the intensity factor with slight severity, we collect $27 \times 689=18,603$ shadowed images with slight intensity while diverse sizes, shapes, and locations. Given different shadow patterns, we analyze shadow and shadow removal effects to image quality and landmark detection performance covering four shadow removal methods, \ie, MaskShadow-GAN \cite{hu2019mask}, SP+M-Net \cite{le2019shadow}, AEFNet \cite{fu2021auto}, and Ours and three landmark detectors, \ie, SAN \cite{dong2018style}, HRNet \cite{wang2020deep}, and LUVLi \cite{kumar2020luvli}. The results are shown in \figref{fig:sub_rm_ld}. 

It turns out that: 1) Shadow affects the image quality and facial landmark detection performance significantly with higher RMSE and NME. More intense the shadow degradation, worse the visual quality and landmark detection performance for all landmark detectors. For example, heavy-intensity shadow images achieve 5.13 NME by landmark detector LUVLi compared to 4.10 NME of slight-intensity shadow images (See \figref{fig:sub_rm_ld}-$\mathcal{C}$(a)). 2) Shadow removal can reduce the performance loss of the image quality and landmark detection caused by shadow. Higher severity achieves higher performance gain after shadow removal. For example, small-size shadow increases the performance loss by 0.3\% after shadow removal by our method via landmark detector HRNet compared to 17.0\% decreasing NME of large-size shadow (See \figref{fig:sub_rm_ld}-$\mathcal{B}$(b)). 3) For all detectors, shadow-removed images by MaskShadow-GAN obtain worse image quality and landmark detection performance due to the introduced artifacts. For example, medium-size shadow-removed images achieve 41.5\% higher RMSE and 4.8\% higher NME by HRNet (See \figref{fig:sub_rm_ld}-$\mathcal{B}$(b)). 4) Intensity, size, and location are dominant factors affecting shadow degradation, \ie, various intensities, sizes, and locations have obviously diverse effects to image quality and facial landmark detection before and after shadow removal. 
For example, \figref{fig:sub_rm_ld}(b) shows that shadow images with large size have higher RMSE (15.22) with shadow-free images compared to that of small size (2.74). However, the relative difference of RMSE of shadow and shadow-free images between different shape severities is within 0.7. The trend is the same as the shadow effect on facial landmark detection performance with various detectors. After shadow removal, image quality and facial landmark detection performance are similar between various shape complexities with comparable RMSEs and NMEs. 

%

\subsection{Implementation Details for Embedding Fusion}

\paragraph{Recovery network ($\tau$).} Given a shadow image $\mathbf{I}$ and a landmark detector $\varphi(\cdot)$, we can obtain ${f}, {h} =\varphi({\mathbf{I}})$, where $f \in$ $\mathds{R}^{N\times C\times H\times W}$ is the detection embedding from landmark detector and $h$ is the output heatmap representing landmark localization. $N, C, H,$ and $W$ are batch size, channel number, height, and width of $f$, respectively. We extract $f$ from the third convolution block of the backbone of landmark detector SAN \cite{dong2018style} with respective field $32\times32$. Then, $f$ is fed into the recovery network $\tau(\cdot)$ for embedding alignment. The architecture of network $\tau$ is listed in \tableref{tab:tau}. The outputs of convolution layers ``Conv1\_2'', ``Conv2\_2'', and ``Conv3\_2'' will be concatenated and go through ``Conv1\_fuse" for the final output. Multi-level feature fusion is designed for less information loss. The output of the recovery network is $\mathbf{T}=\tau(\varphi(\mathbf{I}))$, where  $\mathbf{T}, \in \mathds{R}^{N\times C\times H\times W}$, has the same dimension with $f$.

\setcounter{table}{0}
\begin{table}[ht]
\caption{The architecture of recovery network.}
\centering
\footnotesize
\resizebox{0.95\linewidth}{!}{
\begin{tabular}{r|ccccc}
\toprule
 & Input Channel & Output Channel & Filter Size & Stride & Pad\\
\midrule
Conv1\_1 & 256 & 256 & 3 & 1 & 1\\
Conv1\_2 & 256 & 256 & 3 & 1 & 1\\
Conv2\_1 & 256 & 128 & 3 & 1 & 1\\
Conv2\_2 & 128 & 128 & 3 & 1 & 1\\
Conv3\_1 & 128 & 64 & 3 & 1 & 1\\
Conv3\_2 & 64 & 64 & 3 & 1 & 1\\
Conv\_fuse & 448 & 256 & 1 & 1 & 0\\
\bottomrule
\end{tabular}\label{tab:tau}
}
\end{table}

\paragraph{Mutual attention fusion weights of MAFus.} Given the embedding of shadow removal model $\mathbf{F}=\psi(\mathbf{I})$ and aligned landmark detection embedding $\mathbf{T}=\tau(\varphi(\mathbf{I}))$, we can obtain two similarity results $f^\text{f}(\mathbf{F})$ and $f^\text{t}(\mathbf{T})$ based on non-local module \cite{wang2018non}. Previously, attention for each embedding is calculated based on its similarity result only. In contrast, mutual attention is achieved by a weighted sum of attentions of different modalities. The mutual attentions of shadow removal embedding $\mathbf{A}^\text{f}$ and landmark detection embedding $\mathbf{A}^\text{t}$ are calculated by 
\begin{align}\label{eq:nonlocal-3-supp}
\mathbf{A}^\text{f}(f^\text{f}(\mathbf{F}),f^\text{t}(\mathbf{T}))=\text{softmax}(f^\text{f}(\mathbf{F})+\gamma^\text{t}f^\text{t}(\mathbf{T})),\\
\mathbf{A}^\text{t}(f^\text{f}(\mathbf{F}),f^\text{t}(\mathbf{T}))=\text{softmax}(f^\text{t}(\mathbf{F})+\gamma^\text{f}f^\text{f}(\mathbf{T})),
\end{align}
where $\gamma^\text{t}$ and $\gamma^\text{f}$, $\in \mathds{R}^{N\times 1\times H\times W}$, are pixel-wise attention weights. Both $H$ and $W$ are $32$. $\gamma^\text{t}$ and $\gamma^\text{f}$ are predicted by a convolution layer followed by a BN \cite{ioffe2015batch} layer and the ReLU activation function. The kernel size of convolution layer is 1. The input to the convolution layer is the concatenation of $\mathbf{F}$ and $\mathbf{T}$, \ie, [$\mathbf{F}$, $\mathbf{T}$]. $\mathbf{F}$ has the same dimension with $\mathbf{T}$.

\subsection{Ablation Study on Embedding Fusion}
For the shadow removal embedding $\mathbf{F}=\psi(\mathbf{I})$ and landmark detection embedding $\varphi(\mathbf{I})$, we also verify different embedding fusion strategies built on the baseline method ``AEFNet \cite{fu2021auto} + $\mathcal{L}_\mathrm{det}$". 1) We first directly concatenate the two embeddings, \ie, $\hat{\mathbf{F}} = [\mathbf{F}, \varphi(\mathbf{I})]$, and then feed the $\hat{\mathbf{F}}$ to the decoder of the shadow removal model. 2) We feed the landmark detection embedding $\varphi(\mathbf{I})$ to the recovery network $\tau$ to align the embeddings of $\mathbf{F}$ and $\varphi(\mathbf{I})$ for the whole shadow removal pipeline, and followed by concatenation and shadow removal decoder. The fused embedding $\hat{\mathbf{F}}$ is $[\mathbf{F}, \tau(\varphi(\mathbf{I}))]$. 3) To fully integrate the shadow removal and landmark detection embeddings, we utilize mutual attention to fuse them instead of direct concatenation, \ie, $\hat{\mathbf{F}} = \text{MAFus}(\mathbf{F},\tau(\varphi(\mathbf{I})))$. The shadow removal and landmark detection results are shown in \tableref{tab:abl_fuse}. The landmark detector is SAN.

\setcounter{table}{1}
\begin{table}[ht]
\caption{Ablation study of shadow removal and landmark detection results with different embedding fusion strategies on the $\mathcal{D}_\text{syn}$ dataset.}
\centering
\footnotesize
\resizebox{0.95\linewidth}{!}{
\begin{tabular}{r|ccc|c}
\toprule
\multirow{2}{*}{Methods} & \multicolumn{3}{c|}{Shadow removal / RMSE} & \multirow{2}{2cm}{Landmark detection / NME}\\
\cline{2-4} 
 & Shadow & Non-shadow & All & \\
\midrule
Clean                                & 0.00  & 0.00 & 0.00 & 4.04\\
w/ shadow                            & 33.27 & 0.53 & 8.09 & 5.17  \\
\midrule
Baseline                                            & 7.56 & \color{red}{1.93} & 3.23 & 4.40 \\
w/ [$\mathbf{F}, \varphi(\mathbf{I})$]                  & 7.33 & \color{red}{1.93} & 3.18 & 4.38\\
w/ [$\mathbf{F},\tau(\varphi(\mathbf{I}))$]             & \color{red}{7.15} & 1.96 & \color{red}{3.16} & 4.37\\

w/ $\text{MAFus}(\mathbf{F},\tau(\varphi(\mathbf{I})))$ & 7.16 & 1.96 & \color{red}{3.16} & \color{red}{4.35} \\
\bottomrule

\end{tabular}\label{tab:abl_fuse}
}
\end{table}

It turns out that 1) embedding concatenation of shadow removal model and landmark detection model contributes to shadow removal and landmark detection simultaneously. The landmark detection performance reaches lower NME 4.38 compared to 4.40 NME of the baseline method. The shadow removal performance obtains lower RMSE 3.18 compared to 3.23 RMSE of the baseline method in the whole image. 2) The recovery network $\tau(\cdot)$ can improve the shadow removal and landmark detection performance effectively via embedding alignment. The landmark detection performance reaches 4.37 NME. 3) The mutual attention fusion module further reduces the NME of landmark detection performance to 4.35. It demonstrates that mutual attention can leverage and integrate the features from different networks effectively.


{\small
\bibliographystyle{ieee_fullname}
\bibliography{ref}

\begin{thebibliography}{10}\itemsep=-1pt

\bibitem{belagiannis2017recurrent}
Vasileios Belagiannis and Andrew Zisserman.
\newblock Recurrent human pose estimation.
\newblock In {\em International Conference on Automatic Face \& Gesture
  Recognition (AFGR)}, pages 468--475, 2017.

\bibitem{chen2005estimating}
Yinpeng Chen and Hari Sundaram.
\newblock Estimating complexity of 2d shapes.
\newblock In {\em Workshop on Multimedia Signal Processing (MSPW)}, pages 1--4,
  2005.

\bibitem{tmm21_pasadena}
Yupeng Cheng, Qing Guo, Felix Juefei-Xu, Xiaofei Xie, Shang-Wei Lin, Weisi Lin,
  Wei Feng, and Yang Liu.
\newblock {Pasadena: Perceptually Aware and Stealthy Adversarial Denoise
  Attack}.
\newblock {\em IEEE Transactions on Multimedia (TMM)}, 2021.

\bibitem{cheng2020adversarial}
Yupeng Cheng, Felix Juefei-Xu, Qing Guo, Huazhu Fu, Xiaofei Xie, Shang-Wei Lin,
  Weisi Lin, and Yang Liu.
\newblock Adversarial exposure attack on diabetic retinopathy imagery.
\newblock {\em arXiv preprint arXiv:2009.09231}, 2020.

\bibitem{dong2018style}
Xuanyi Dong, Yan Yan, Wanli Ouyang, and Yi Yang.
\newblock Style aggregated network for facial landmark detection.
\newblock In {\em Conference on Computer Vision and Pattern Recognition
  (CVPR)}, pages 379--388, 2018.

\bibitem{fu2021let}
Lan Fu, Hongkai Yu, Felix Juefei-Xu, Jinlong Li, Qing Guo, and Song Wang.
\newblock Let there be light: Improved traffic surveillance via detail
  preserving night-to-day transfer.
\newblock {\em IEEE Transactions on Circuits and Systems for Video Technology
  (TCSVT)}, 2021.

\bibitem{fu2021auto}
Lan Fu, Changqing Zhou, Qing Guo, Felix Juefei-Xu, Hongkai Yu, Wei Feng, Yang
  Liu, and Song Wang.
\newblock Auto-exposure fusion for single-image shadow removal.
\newblock In {\em Conference on Computer Vision and Pattern Recognition
  (CVPR)}, pages 10571--10580, 2021.

\bibitem{arxiv21_advhaze}
Ruijun Gao, Qing Guo, Felix Juefei-Xu, Hongkai Yu, and Wei Feng.
\newblock {AdvHaze: Adversarial Haze Attack}.
\newblock {\em arXiv preprint arXiv:2104.13673}, 2021.

\bibitem{gao2020making}
Ruijun Gao, Qing Guo, Felix Juefei-Xu, Hongkai Yu, Xuhong Ren, Wei Feng, and
  Song Wang.
\newblock Making images undiscoverable from co-saliency detection.
\newblock {\em arXiv preprint arXiv:2009.09258}, 2020.

\bibitem{arxiv21_ara}
Ruijun Gao, Qing Guo, Qian Zhang, Felix Juefei-Xu, Hongkai Yu, and Wei Feng.
\newblock {Adversarial Relighting against Face Recognition}.
\newblock {\em arXiv preprint arXiv:2108.07920}, 2021.

\bibitem{guo2020towards}
Jianzhu Guo, Xiangyu Zhu, Yang Yang, Fan Yang, Zhen Lei, and Stan~Z Li.
\newblock Towards fast, accurate and stable 3d dense face alignment.
\newblock In {\em European Conference on Computer Vision (ECCV)}, pages
  152--168, 2020.

\bibitem{iccv21_advmot}
Qing Guo, Ziyi Cheng, Felix Juefei-Xu, Lei Ma, Xiaofei Xie, Yang Liu, and
  Jianjun Zhao.
\newblock {Learning to Adversarially Blur Visual Object Tracking}.
\newblock In {\em Proceedings of the IEEE International Conference on Computer
  Vision (ICCV)}. IEEE, October 2021.

\bibitem{neurips20_abba}
Qing Guo, Felix Juefei-Xu, Xiaofei Xie, Lei Ma, Jian Wang, Bing Yu, Wei Feng,
  and Yang Liu.
\newblock {Watch out! Motion is Blurring the Vision of Your Deep Neural
  Networks}.
\newblock In {\em Advances in Neural Information Processing Systems (NeurIPS)},
  2020.

\bibitem{eccv20_spark}
Qing Guo, Xiaofei Xie, Felix Juefei-Xu, Lei Ma, Zhongguo Li, Wanli Xue, Wei
  Feng, and Yang Liu.
\newblock {SPARK: Spatial-aware Online Incremental Attack Against Visual
  Tracking}.
\newblock In {\em European Conference on Computer Vision (ECCV)}, Aug 2020.

\bibitem{hanrahan1993reflection}
Pat Hanrahan and Wolfgang Krueger.
\newblock Reflection from layered surfaces due to subsurface scattering.
\newblock In {\em Conference on Computer Graphics and Interactive Techniques
  (CGIT)}, pages 165--174, 1993.

\bibitem{hnewa2020object}
Mazin Hnewa and Hayder Radha.
\newblock Object detection under rainy conditions for autonomous vehicles: A
  review of state-of-the-art and emerging techniques.
\newblock {\em IEEE Signal Processing Magazine (SPM)}, 38(1):53--67, 2020.

\bibitem{hu2019mask}
Xiaowei Hu, Yitong Jiang, Chi-Wing Fu, and Pheng-Ann Heng.
\newblock Mask-shadowgan: Learning to remove shadows from unpaired data.
\newblock In {\em International Conference on Computer Vision (ICCV)}, pages
  2472--2481, 2019.

\bibitem{arxiv21_advbokeh}
Yihao Huang, Felix Juefei-Xu, Qing Guo, Weikai Miao, Yang Liu, and Geguang Pu.
\newblock Advbokeh: Learning to adversarially defocus blur.
\newblock {\em arXiv preprint}, 2021.

\bibitem{inoue2020learning}
Naoto Inoue and Toshihiko Yamasaki.
\newblock Learning from synthetic shadows for shadow detection and removal.
\newblock {\em IEEE Transactions on Circuits and Systems for Video Technology
  (TCSVT)}, 2020.

\bibitem{ioffe2015batch}
Sergey Ioffe and Christian Szegedy.
\newblock Batch normalization: Accelerating deep network training by reducing
  internal covariate shift.
\newblock In {\em International Conference on Machine Learning (ICML)}, pages
  448--456, 2015.

\bibitem{johnson2016perceptual}
Justin Johnson, Alexandre Alahi, and Li Fei-Fei.
\newblock Perceptual losses for real-time style transfer and super-resolution.
\newblock In {\em European Conference on Computer Vision (ECCV)}, pages
  694--711, 2016.

\bibitem{juefei2013image}
Felix Juefei-Xu and Marios Savvides.
\newblock An image statistics approach towards efficient and robust refinement
  for landmarks on facial boundary.
\newblock In {\em International Conference on Biometrics: Theory, Applications
  and Systems (BTAS)}, pages 1--8, 2013.

\bibitem{kumar2020luvli}
Abhinav Kumar, Tim~K Marks, Wenxuan Mou, Ye Wang, Michael Jones, Anoop Cherian,
  Toshiaki Koike-Akino, Xiaoming Liu, and Chen Feng.
\newblock Luvli face alignment: Estimating landmarks' location, uncertainty,
  and visibility likelihood.
\newblock In {\em Conference on Computer Vision and Pattern Recognition
  (CVPR)}, pages 8236--8246, 2020.

\bibitem{le2019shadow}
Hieu Le and Dimitris Samaras.
\newblock Shadow removal via shadow image decomposition.
\newblock In {\em International Conference on Computer Vision (ICCV)}, pages
  8578--8587, 2019.

\bibitem{li2020structured}
Weijian Li, Yuhang Lu, Kang Zheng, Haofu Liao, Chihung Lin, Jiebo Luo, Chi-Tung
  Cheng, Jing Xiao, Le Lu, Chang-Fu Kuo, et~al.
\newblock Structured landmark detection via topology-adapting deep graph
  learning.
\newblock In {\em European Conference on Computer Vision (ECCV)}, pages
  266--283, 2020.

\bibitem{iccv21_flat}
Yiming Li, Congcong Wen, Felix Juefei-Xu, and Chen Feng.
\newblock {Fooling LiDAR Perception via Adversarial Trajectory Perturbation}.
\newblock In {\em Proceedings of the IEEE International Conference on Computer
  Vision (ICCV)}. IEEE, October 2021.

\bibitem{li2020celeb}
Yuezun Li, Xin Yang, Pu Sun, Honggang Qi, and Siwei Lyu.
\newblock Celeb-df: A large-scale challenging dataset for deepfake forensics.
\newblock In {\em Conference on Computer Vision and Pattern Recognition
  (CVPR)}, pages 3207--3216, 2020.

\bibitem{liu2016joint}
Feng Liu, Dan Zeng, Qijun Zhao, and Xiaoming Liu.
\newblock Joint face alignment and 3d face reconstruction.
\newblock In {\em European Conference on Computer Vision (ECCV)}, pages
  545--560, 2016.

\bibitem{liu2020learning}
Nian Liu, Ni Zhang, and Junwei Han.
\newblock Learning selective self-mutual attention for rgb-d saliency
  detection.
\newblock In {\em Conference on Computer Vision and Pattern Recognition
  (CVPR)}, pages 13756--13765, 2020.

\bibitem{liu2018exploring}
Yu Liu, Fangyin Wei, Jing Shao, Lu Sheng, Junjie Yan, and Xiaogang Wang.
\newblock Exploring disentangled feature representation beyond face
  identification.
\newblock In {\em Conference on Computer Vision and Pattern Recognition
  (CVPR)}, pages 2080--2089, 2018.

\bibitem{pei2019effects}
Yanting Pei, Yaping Huang, Qi Zou, Xingyuan Zhang, and Song Wang.
\newblock Effects of image degradation and degradation removal to cnn-based
  image classification.
\newblock {\em IEEE Transactions on Pattern Analysis and Machine Intelligence
  (TPAMI)}, 2019.

\bibitem{qu2017deshadownet}
Liangqiong Qu, Jiandong Tian, Shengfeng He, Yandong Tang, and Rynson~WH Lau.
\newblock Deshadownet: A multi-context embedding deep network for shadow
  removal.
\newblock In {\em Conference on Computer Vision and Pattern Recognition
  (CVPR)}, pages 4067--4075, 2017.

\bibitem{sagonas2013300}
Christos Sagonas, Georgios Tzimiropoulos, Stefanos Zafeiriou, and Maja Pantic.
\newblock 300 faces in-the-wild challenge: The first facial landmark
  localization challenge.
\newblock In {\em International Conference on Computer Vision Workshops
  (ICCVW)}, pages 397--403, 2013.

\bibitem{shor2008the}
Yael Shor and Dani Lischinski.
\newblock The shadow meets the mask: Pyramid-based shadow removal.
\newblock {\em Computer Graphics Forum (CGF)}, 27(2):577--586, 2008.

\bibitem{sun2018feature}
Zhun Sun, Mete Ozay, Yan Zhang, Xing Liu, and Takayuki Okatani.
\newblock Feature quantization for defending against distortion of images.
\newblock In {\em Conference on Computer Vision and Pattern Recognition
  (CVPR)}, pages 7957--7966, 2018.

\bibitem{thies2016face2face}
Justus Thies, Michael Zollhofer, Marc Stamminger, Christian Theobalt, and
  Matthias Nie{\ss}ner.
\newblock Face2face: Real-time face capture and reenactment of rgb videos.
\newblock In {\em Conference on Computer Vision and Pattern Recognition
  (CVPR)}, pages 2387--2395, 2016.

\bibitem{icme21_xray}
Binyu Tian, Qing Guo, Felix Juefei-Xu, Wen~Le Chan, Yupeng Cheng, Xiaohong Li,
  Xiaofei Xie, and Shengchao Qin.
\newblock {Bias Field Poses a Threat to DNN-Based X-Ray Recognition}.
\newblock In {\em IEEE International Conference on Multimedia and Expo (ICME)},
  2021.

\bibitem{ijcai21_ava}
Binyu Tian, Felix Juefei-Xu, Qing Guo, Xiaofei Xie, Xiaohong Li, and Yang Liu.
\newblock {AVA: Adversarial Vignetting Attack against Visual Recognition}.
\newblock In {\em Proceedings of the International Joint Conference on
  Artificial Intelligence (IJCAI)}, 2021.

\bibitem{toshev2014deeppose}
Alexander Toshev and Christian Szegedy.
\newblock Deeppose: Human pose estimation via deep neural networks.
\newblock In {\em Conference on Computer Vision and Pattern Recognition
  (CVPR)}, pages 1653--1660, 2014.

\bibitem{valle2018deeply}
Roberto Valle, Jose~M Buenaposada, Antonio Valdes, and Luis Baumela.
\newblock A deeply-initialized coarse-to-fine ensemble of regression trees for
  face alignment.
\newblock In {\em European Conference on Computer Vision (ECCV)}, pages
  585--601, 2018.

\bibitem{wang2018stacked}
Jifeng Wang, Xiang Li, and Jian Yang.
\newblock Stacked conditional generative adversarial networks for jointly
  learning shadow detection and shadow removal.
\newblock In {\em Conference on Computer Vision and Pattern Recognition
  (CVPR)}, pages 1788--1797, 2018.

\bibitem{wang2020deep}
Jingdong Wang, Ke Sun, Tianheng Cheng, Borui Jiang, Chaorui Deng, Yang Zhao,
  Dong Liu, Yadong Mu, Mingkui Tan, Xinggang Wang, et~al.
\newblock Deep high-resolution representation learning for visual recognition.
\newblock {\em IEEE Transactions on Pattern Analysis and Machine Intelligence
  (TPAMI)}, 2020.

\bibitem{acmmm20_amora}
Run Wang, Felix Juefei-Xu, Qing Guo, Yihao Huang, Xiaofei Xie, Lei Ma, and Yang
  Liu.
\newblock {Amora: Black-box Adversarial Morphing Attack}.
\newblock In {\em Proceedings of the ACM International Conference on Multimedia
  (ACM MM)}, 2020.

\bibitem{wang2018non}
Xiaolong Wang, Ross Girshick, Abhinav Gupta, and Kaiming He.
\newblock Non-local neural networks.
\newblock In {\em Conference on Computer Vision and Pattern Recognition
  (CVPR)}, pages 7794--7803, 2018.

\bibitem{wu2018look}
Wayne Wu, Chen Qian, Shuo Yang, Quan Wang, Yici Cai, and Qiang Zhou.
\newblock Look at boundary: A boundary-aware face alignment algorithm.
\newblock In {\em Conference on Computer Vision and Pattern Recognition
  (CVPR)}, pages 2129--2138, 2018.

\bibitem{wu2019facial}
Yue Wu and Qiang Ji.
\newblock Facial landmark detection: A literature survey.
\newblock {\em International Journal of Computer Vision (IJCV)},
  127(2):115--142, 2019.

\bibitem{zhai2020s}
Liming Zhai, Felix Juefei-Xu, Qing Guo, Xiaofei Xie, Lei Ma, Wei Feng,
  Shengchao Qin, and Yang Liu.
\newblock It's raining cats or dogs? adversarial rain attack on dnn perception.
\newblock {\em arXiv preprint arXiv:2009.09205}, 2020.

\bibitem{zhang2020freenet}
Jiangning Zhang, Xianfang Zeng, Mengmeng Wang, Yusu Pan, Liang Liu, Yong Liu,
  Yu Ding, and Changjie Fan.
\newblock Freenet: Multi-identity face reenactment.
\newblock In {\em Conference on Computer Vision and Pattern Recognition
  (CVPR)}, pages 5326--5335, 2020.

\bibitem{zhang2020portrait}
Xuaner Zhang, Jonathan~T Barron, Yun-Ta Tsai, Rohit Pandey, Xiuming Zhang, Ren
  Ng, and David~E Jacobs.
\newblock Portrait shadow manipulation.
\newblock {\em ACM Transactions on Graphics (TOG)}, 39(4):78--1, 2020.

\bibitem{zhang2014facial}
Zhanpeng Zhang, Ping Luo, Chen~Change Loy, and Xiaoou Tang.
\newblock Facial landmark detection by deep multi-task learning.
\newblock In {\em European Conference on Computer Vision (ECCV)}, pages
  94--108. Springer, 2014.

\bibitem{zhao2021learning}
Tianchen Zhao, Xiang Xu, Mingze Xu, Hui Ding, Yuanjun Xiong, and Wei Xia.
\newblock Learning self-consistency for deepfake detection.
\newblock In {\em International Conference on Computer Vision (ICCV)}, pages
  15023--15033, 2021.

\bibitem{zhu2018bidirectional}
Lei Zhu, Zijun Deng, Xiaowei Hu, Chi-Wing Fu, Xuemiao Xu, Jing Qin, and
  Pheng-Ann Heng.
\newblock Bidirectional feature pyramid network with recurrent attention
  residual modules for shadow detection.
\newblock In {\em European Conference on Computer Vision (ECCV)}, pages
  121--136, 2018.

\bibitem{zhu2015high}
Xiangyu Zhu, Zhen Lei, Junjie Yan, Dong Yi, and Stan~Z Li.
\newblock High-fidelity pose and expression normalization for face recognition
  in the wild.
\newblock In {\em Conference on Computer Vision and Pattern Recognition
  (CVPR)}, pages 787--796, 2015.

\bibitem{zou2019learning}
Xu Zou, Sheng Zhong, Luxin Yan, Xiangyun Zhao, Jiahuan Zhou, and Ying Wu.
\newblock Learning robust facial landmark detection via hierarchical structured
  ensemble.
\newblock In {\em International Conference on Computer Vision (ICCV)}, pages
  141--150, 2019.

\end{thebibliography}
}

\newpage
\thispagestyle{empty} 

\setcounter{figure}{0}    
\begin{figure*}[ht]
\centering
\includegraphics[width=1.0\textwidth]{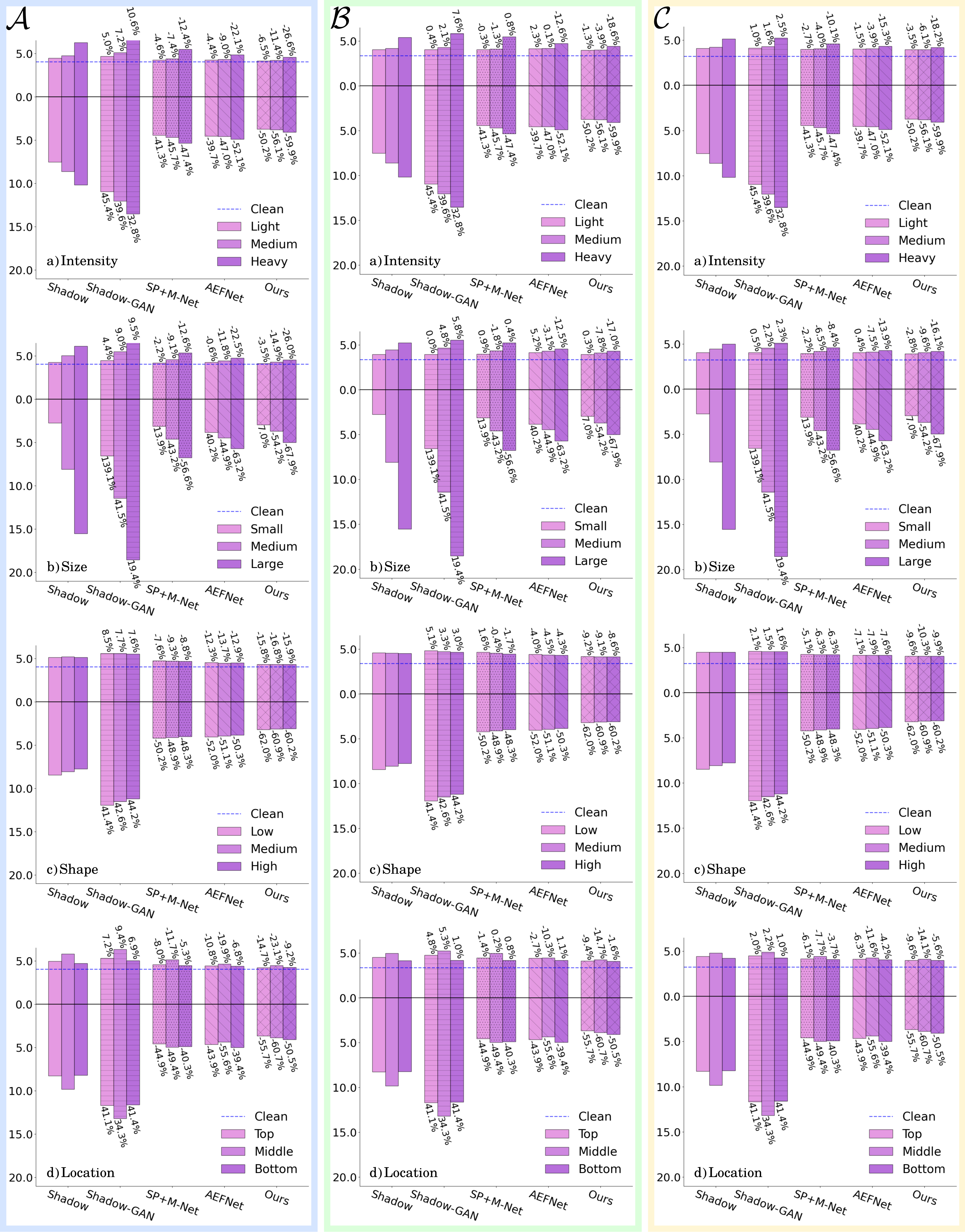}
\caption{Shadow pattern analysis of shadow removal and landmark detection performance on $\mathcal{D}_\text{syn}$. ($\mathcal{A}$-$\mathcal{C}$): shadow removal (RMSE) and landmark detection (NME) results with shadow removal methods (\ie, MaskShadow-GAN \cite{hu2019mask}, SP+M-Net \cite{le2019shadow}, AEFNet \cite{fu2021auto}, and Ours) and detectors (\ie, SAN \cite{dong2018style} ($\mathcal{A}$), HRNet \cite{wang2020deep} ($\mathcal{B}$), and LUVLi \cite{kumar2020luvli} ($\mathcal{C}$)). (a-d): landmark detection (NME) and shadow removal (RMSE) results of $\mathcal{D}_\text{syn}$ for \textbf{intensity} (a), \textbf{size} (b), \textbf{shape} (c), and \textbf{location} (d). Blue dash line represents the result on clean images by the pre-trained landmark detectors.
Each group along the x-axis represents results on shadow images (\ie, Shadow), and shadow-removed images with four shadow removal methods (\eg, MaskShadow-GAN/SP+M-Net/AEFNet/Ours). Each color represents a severity type. Relative performance gains, \ie, the percent of NME/RMSE drops, after shadow removal compared to shadow images are listed for MaskShadow-GAN, SP+M-Net, AEFNet, and Ours. Note: Shadow-GAN denotes the MaskShadow-GAN.    
}
\label{fig:sub_rm_ld}
\end{figure*}


\end{document}